\documentclass[letterpaper,twocolumn,fleqn]{article} 

\usepackage{ist}

\usepackage{cite}
\usepackage{array}
\usepackage{graphicx}
\usepackage{fix-cm}
\usepackage{times}
\usepackage{dirtytalk}
\usepackage{textcomp}
\usepackage{eurosym}
\usepackage{hyperref}
\usepackage{nomencl,etoolbox,ragged2e,siunitx}
\usepackage[utf8]{inputenc}
\usepackage[english]{babel}
\usepackage{enumitem}
\usepackage{multicol}
\usepackage{multirow}
\usepackage{booktabs}

\title{Detecting Deepfake Videos Using Euler Video Magnification}
\author{Rashmiranjan Das~$^1$,
Gaurav Negi~$^1$ and Alan F. Smeaton~$^{1,2}$\\
$^1$School of Computing and~ $^2$Insight Centre for Data Analytics\\
Dublin City University, Glasnevin, Dublin, 9, Ireland.\\
Email: alan.smeaton@dcu.ie} 

\date{} 

\begin{document}
\maketitle
\thispagestyle{empty} 

\begin{abstract} 

Recent advances in  artificial intelligence make it progressively hard to distinguish between genuine and counterfeit media, especially images and videos. One  recent development is the rise of deepfake videos,  based on manipulating videos using advanced machine learning techniques. This involves replacing the face of an individual from a source video with the face of a second person, in the destination video. This  idea is becoming progressively refined as deepfakes are getting progressively seamless and simpler to compute. Combined with the outreach and speed of social media, deepfakes could easily fool individuals when depicting someone saying things that never happened and thus could persuade people in believing fictional scenarios, creating distress, and spreading fake news. In this paper, we examine a technique for possible identification of deepfake videos. We use Euler video magnification which applies spatial decomposition and temporal filtering on video data to highlight and magnify hidden features like skin pulsation and subtle motions. Our approach uses features extracted from the Euler technique to train three models to classify counterfeit and unaltered videos and compare the results with existing techniques.  
\end{abstract}


\section{Introduction}
Deepfakes  describes a technique for artificially manipulating video, initially applied to swap celebrities faces into video recordings which were shared on sites like Reddit  \cite{koopman2018detection}. They operate by replacing the face of one person in an original video, with a second person, inserted so that head movement, facial expressions, lighting and lip syncing when talking, are all exactly as in the original video.   While many thousands of images of the second person to be superimposed into the deepfake, are usually required, recent work has shown that good deepfakes can be generated with a reduced number \cite{singh2020using}. This means  fake videos can be  realistic  and can be generated with small amounts of training data. 

Deepfakes can be recognised as both an opportunity and a threat because they permit users with relatively little computing experience in machine learning or computer programming to generate almost  seamless fake videos. The availability of state of the art deep learning libraries such as TensorFlow \cite{abadi2016tensorflow} and Keras \cite{chollet2015keras}, with enough accessible training data of facial images, allows generation of fake video recordings whose quality is so good they can be very persuading \cite{guera2018deepfake}.

The initial implementations of deepfakes relied on convolutional autoencoders \cite{thies2016face2face}. Images of both subjects are reduced to lower dimensions using an encoder and reconstructed using a decoder. This training is performed for both source and destination facial expressions. In order to perform a face swap, a trained encoder of the source is mapped with a decoder trained on the target subject's face. An upgrade to this technique is by adding a generative adversarial network (GAN) in the decoder \cite{goodfellow2016nips,arjovsky2017wasserstein}. GANs consist of two modules, a generator and a discriminator. The task of the generator is to develop images resembling the source while the discriminator determines if the image is counterfeited. It is an iterative process, which makes deepfakes realistic as they are constantly learning.

The availability of such sophisticated techniques for deepfake generation in the hands of ordinary researchers and their possible exploitation by other persons have escalated concerns about their possible misuse. Applications such as Deepfacelab \cite{petrov2020deepfacelab}, FakeApp and OpenFaceSwap are GUI based tools made accessible to relatively untrained researchers to create deepfake videos. With these tools, it becomes progressively possible for video evidence to be altered for political tension, false video evidence and fake news. Hence, this poses a challenge for society as well as an opportunity for creating novel entertainment, but it  demands an effective technique for the detection of such counterfeit video.

\section{Related Work on Deepfake Detection}

One approach to deepfake detection focuses on psychological signals in the video \cite{li2018ictu},  proposing a detection method  by observing eye blinking in videos, a psychological signal not well presented in synthesised videos. This  is based on a novel deep learning model combining a convolutional neural network (CNN) with a recursive neural network (RNN) that captures phenomenological and temporal regularities in the eye blinking process. Since the training images used to generate deep fakes  do not usually include images of the subject with eyes closed, this is a clever approach, though it  can be circumvented by intentionally integrating images into the training data with eyes closed. 

The work in \cite{li2018detection} exploits  colour disparity between GAN-generated images and real images in the non-RGB colour space to classify them. Again in the work reported in \cite{mccloskey2018detecting}  analysed the colour difference between GAN images and real images. However, it is not clear if this approach can be applied to inspecting local areas in the image, as would be needed in the deepFake case.

The paper which is most similar to our work is by Fernandes {\em et al.} \cite{fernandes2019predicting}, where they investigate the heart rates of people in deepfake videos and real videos using Neural ODE. The objective of this paper is to generate a heart rate from deepfake videos, which are assumed to have no heartbeat. However, this approach may not perform very well in case of detecting deepfakes as deepfakes  are usually not stable and are not under perfect lighting conditions, so it can be difficult to obtain a stable heartbeat.

\section{Euler Video Magnification}
Eulerian Video Magnification (EVM) \cite{wu2012eulerian} is a technique which  can uncover fleeting and hidden details in videos that will, in general, be hard to see with the human eye. EVM  magnifies and visualises temporal variations in spatial and/or colour aspects in videos. The method emphasises subtle  changes which occur naturally and are encoded within the video but  not seen when viewed. For instance, one can enhance the slight colour changes in videos which include exposed human skin such as around the face where the capillaries in the skin show pulsation due to the bloodstream and blood flow changes caused, in turn from heart rate. 
This is similar to Differential Imaging Forensics introduced in \cite{bourquard2019differential} which also reveal  latent visual features in videos that are not perceivable by human observers.

The basis for EVM is that these difficult-to-see changes happen at specific  frequencies that we can expand using a static window in the frequency space. For instance, in a plucked guitar, each guitar string resonates at a different frequency, so to amplify a string's vibration EVM looks at  pixel variations in the respective frequencies of the note being played.  Similarly to amplifying human pulse visualisation in exposed skin  around the face, one can consider pixel changes in frequencies somewhere in the range 1.0 to 3Hz and  process this as it resonates to between 60 and 180 beats per minute.

\begin{figure}[htp]
    \centering
    \includegraphics[width=9cm]{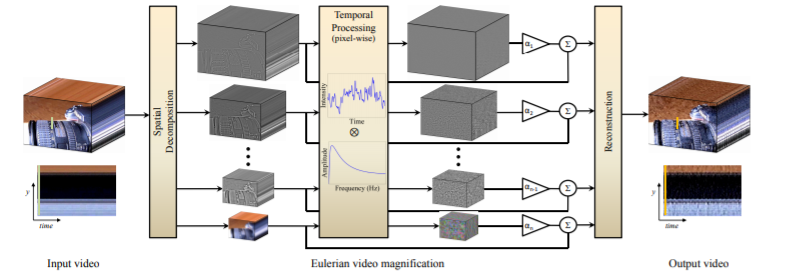}
    \caption{Euler magnification architecture taken from \cite{wu2012eulerian}}
    \label{fig:Euler magnification technique}
\end{figure} \leavevmode\newline

\noindent 
The Eulerian amplification process consists of steps shown in Figure~\ref{fig:Euler magnification technique}. A video is decomposed into images  broken down using a Laplacian pyramid into various frequency ranges. The temporal changes concerning pixels in all frequency ranges of the Laplacian pyramid are bandpass filtered to select important and relevant frequency bands which are amplified by a magnification factor and this outcome is added to the respective signal. Amplified signals which belong to different frequency bands in the Laplacian pyramid are flattened to generate the last yield. The key attribute is the temporal frequency band which can be specified by adjusting the high and low cutoff frequencies for the filter.
One way to consider this is as continuously stacking small variants of the picture on top of each other. This procedure makes a pyramid shape with the base as the first picture and gradually compressing as the pyramid rises.

When we extract the frequency band of interest we amplify the signal and add it back to the source data. 
Adjusting the amplification factor of the bandpass signal results in a larger boost to the temporal bandpass. Changing these parameters can make variations in the scene more apparent but large amplification can add artefacts to the result. 

\section{Video Data  For Deepfake Detection}
\label{sec:data}

During 2020, AWS, Facebook, Microsoft, and others joined  together to build and run a Deepfake Detection Challenge, offering a prize fund of US\$1M to researchers   taking part. The Deepfake Detection Challenge Dataset (DFDC) is described in \cite{dolhansky2019deepfake}. The challenge was hosted on the Kaggle website and 2,265 teams took part in the activity, making more than 3,000 submissions.

The full DFDC dataset consists of 124,000 videos, some of which are real, some are deepfakes.  Subjects in the videos are from varied ethnicities and have different skin tones, genders, lighting conditions and head poses, and activities.
DFDC deepfakes were generated using the whole range of manipulation techniques, such tampering with the intent of representing the real adversarial space of facial manipulation, though no further details of  methods used were provided to  participants in the challenge.

All video clips in the DFDC training set were left at their original resolution and quality, so deriving appropriate augmentations of the training set was left as an exercise to  researchers.  

When the DFDC was complete and results processed, the best system achieved  82.56\% accuracy, based on an ensemble of techniques. This is  important  for two reasons. First,  because there are many ways to generate deepfakes, these need to  be counteracted by using many ways to detect deepfakes. Secondly,  because an individual  generation technique may  require different detection techniques, depending on the video that is generated, thus an ensemble of  detection techniques, is appropriate.

Along with the  DFDC data set, we generated our own set of deepfake videos, created using the Deepfacelab application.  Participants' consent was given and participants signed an agreement approved by the DCU School of Computing Research Ethics Committee. A group of 30 participants
each submitted  a video of 10 seconds  recorded in a controlled environment with suitable lighting at 1080p resolution and the participants remained quite still unlike some of the DFDC videos. Several face swaps were performed using the H64, H128 and SAE techniques \cite{maksutov2020methods}. 

These are autoencoder techniques that reduce the data to smaller dimensions, for example, the H64 model compresses the data into 64x64 pixels. Each face swap video was trained for 30,000 epochs with a mean loss of 0.0630. All these videos are of the same dimension as in DFDC dataset with 30 frames per second. The duration of the videos is 10 secs. A labelled data set was then created by merging the above data sets and this was also used in our experiments.
All  processing was done on an IdeaPad L340 with an NVidia graphic card GTX 1650 and 8 cores, Cuda enabled version 10.1.

\section{Methodology}

This paper explores the use of  Euler Video Magnification as a way to pre-process video and use the result as an indicator of whether a video is a deepfake or not. We 
perform both EVM-based colour and movement amplification on  videos to explore if the resulting differences can distinguish between original and deepfake videos. 

A related technique to our work on colour-based EVM is photoplethysmography (PPG) \cite{mousavi2019blood}. This is a process to identify fluctuations in blood volume by shining  light of a given wavelength onto the skin and  measuring changes in light assimilation. The pumping of heart drives blood to the skin surface in an oscillation cycle and it is the differences in the colour of oxygenated blood that causes changes in light assimilation which is thus a measure of heart rate. Photoplethysmography can also be used to recognise human activities
\cite{brophy2018interpretable} but in this paper we are interested in using colour-based EVM on exposed skin areas such as the face, to see if a pulse can be detected.  Such a process is robust to different skin tones and small motion of the subject and our interest is to see if such EVM-based pulses are present in deepfakes as well as in original videos.

Along with measuring minute colour changes, EVM can also  magnify tiny motions by a subject in a video. One of the characteristics of people  is natural tremor. This is a naturally occurring oscillatory motion which is frequent but not observable to the naked eye due to its very small amplitude. Their recurrence is within the scope of 8 to 12 Hz. These periodic motions have been observed to stay with age and are caused by  compressions that are caused in muscles of the limbs.  

\begin{figure}[htp]
    \centering
    \includegraphics[width=0.8\columnwidth]{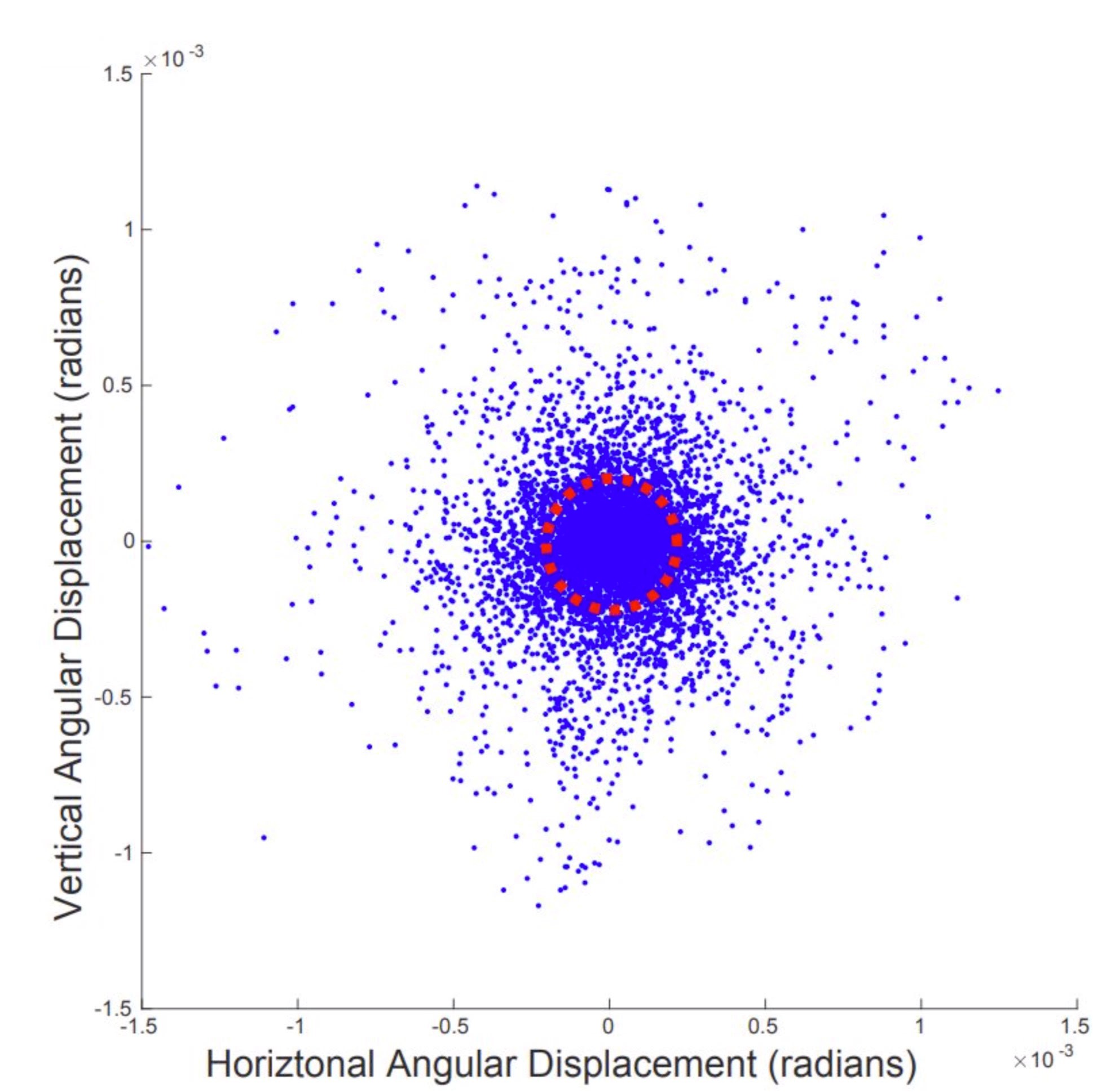}
    \caption{Horizontal and vertical displacement of camera shots taken in burst mode, caused by natural tremors, taken from \cite{wronski2019handheld}.
    \label{fig:Natural-tremors}}
\end{figure}

To illustrate this, Figure~\ref{fig:Natural-tremors} taken from \cite{wronski2019handheld} displays vertical and horizontal displacements that occur naturally to a participant while taking a series of burst shot images with a smartphone camera, those displacements being caused by natural human tremors. The graph was based on 86 burst shots  and the circle formed with red dots marks one standard deviation of movement in any direction. The observation we make from the graph is that human tremors are symmetrically distributed across all directions.  Such small motions  due to natural tremors could be magnified by Euler magnification so that they may be detectable as a differentiator between real and deepfake videos.

We processed videos with EVM, both deepfakes and originals, and then subjected them to three different techniques to extract  features which we used for video classification.

\subsection{Technique 1: SSIM}
The first of the techniques we use is SSIM which is a comparison technique used to compare two frames, evaluate their likeness and calculate a similarity index for the video based on visual structures. The Structural Similarity Index (SSIM) is a perceptual metric that quantifies image quality degradation caused by processing such as data compression or by losses in data transmission, and for deepfake videos, a quality degradation in frames are due to a less well-trained neural net. It is a full reference metric that requires two images from similar image capture.

\begin{figure}[htp]
    \centering
    \includegraphics[width=\columnwidth]{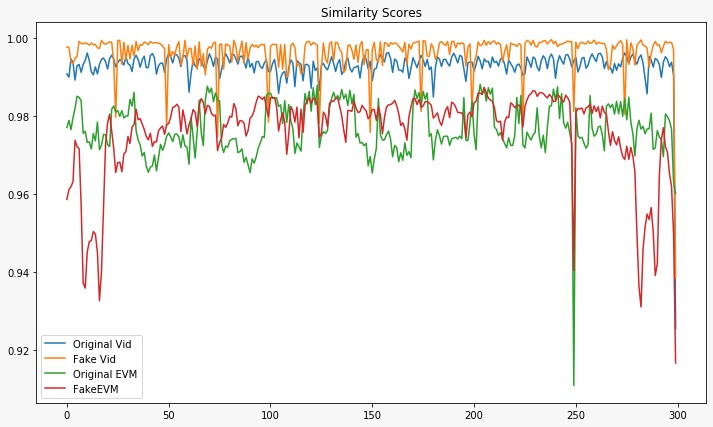}
    \caption{Inter-frame dissimilarity for original, deepfake, Euler magnified original and Euler magnified deepfake videos.
    \label{fig:Similarity plot}}
\end{figure}

\begin{figure}[htp]
    \centering
    \includegraphics[width=6cm]{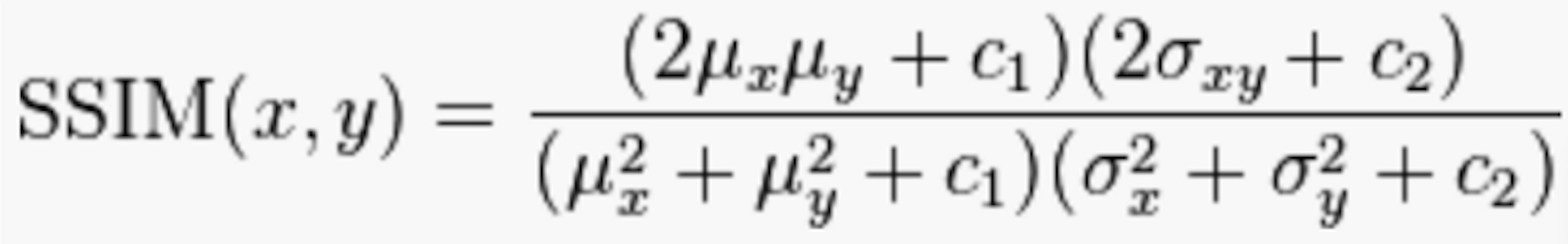}
    \caption{Structural similarity index of two windows x and y of common size NxN. $\mu$ is the average value of (x and y).  $\sigma^{2}$ is the variance. $\sigma$ is the covariance of x and y. c1 and c2 two variables to stabilise the division with weak denominator. }
    \label{fig:Similarity index}
\end{figure}

SSIM is calculated based on  luminance, contrast and structure. Comparing the time series of SSIM of an original video and its EVM equivalent in Figure~\ref{fig:Similarity plot}  it can be seen that Euler magnification enhances  inconsistencies among adjacent  frames of deepfake videos. 
Such inconsistencies in frames are due to pixelated faces and such irregularities will have been intensified and magnified as a result of  Euler magnification where the spatial amplification factor of EVM magnifies the irregularity and hence there are drops in the similarity index.
SSIM is defined in Figure~\ref{fig:Similarity index} \cite{wang2003multiscale}.

\subsection{Technique 2: LSTM}
The idea of using a long short term memory (LSTM) network within a neural network architecture is to help the model learn long term dependencies across the data series. LSTM networks were first presented by Hochreiter \& Schmidhuber in 1997 \cite{hochreiter1997lstm} and their original idea has been upgraded numerous times. The LSTM model's primary objective is to recollect information over an extensive stretch of time. Unlike an RNN with its single tanh layer, LSTMs have four strategically arranged modules. On this basis, LSTM has been used by us for our classification task.

The second technique we use builds on the amount of success achieved by CNN models in video analysis, by adding a Long Short Term Memory (LSTM) network into a neural network architecture. This  can be used to learn any long term dependencies in a data sequence. The LSTM is coupled with the inception module to learn discriminative features from video frames \cite{kumar2020detecting}.

Inception V3 includes an Inception module where there are changes in the spatial convolutions to depth-wise separable convolutions.  Our model is build using CNN network layers for feature extraction followed by an LSTM layer for temporal analysis of Euler magnified videos. Our network has fully connected layers and a dropout layer to make sure that there is no over-fitting. The total number of trainable parameters used is 5,500,898 and these are used as input for the LSTM network and 2 node network working as a detector for deepfake videos from original videos. 

To obtain  ground truth, the neural network was trained on videos which were not modified by the Euler magnified method. The hidden layer had a `relu' activation function, while the last layer had `softmax' as the activation function. We calculated the loss and accuracy of the technique on both the training and test sets. We performed the same sequential steps on the same sets of the video but on the Euler magnified form of the video for comparison on how the technique compares to the standard classification techniques. We used an LSTM network with 512 widths and dropout of 0.5 to randomly set values of outgoing edges of hidden layers to zero. The last layer is constructed using a softmax activation layer to predict video class.

\subsection{Technique 3: Heart Rate Estimation}

\begin{figure}[htp]
    \centering
    \includegraphics[width=\columnwidth]{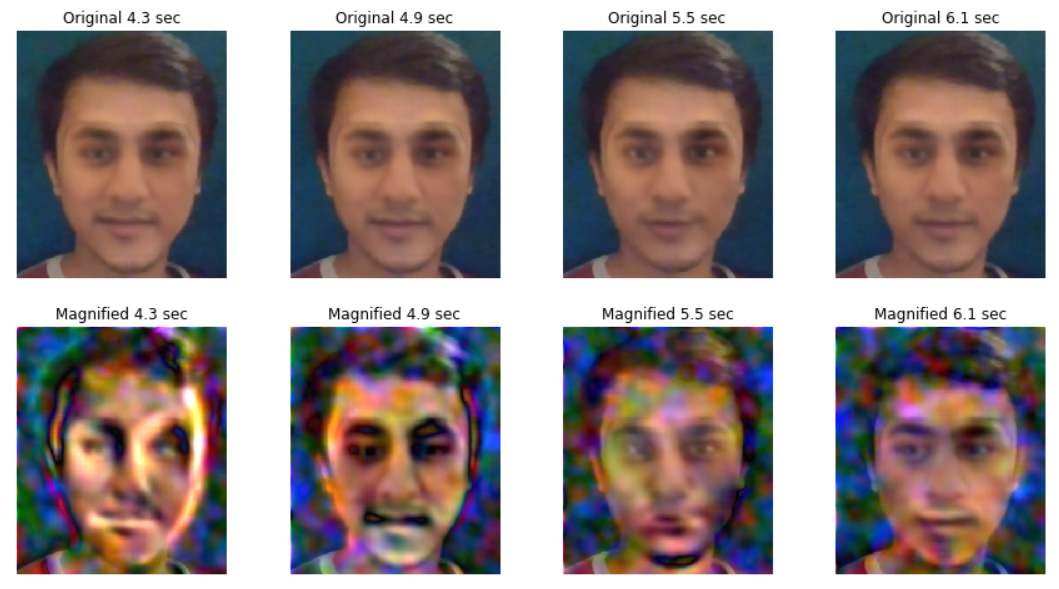}
    \caption{EVM on deepfake video}
    \label{fig:gaurav EVM}
\end{figure}

When computing Euler magnification on a deepfake video, we observed that deepfake videos also exhibit pulsation as seen in  Figure~\ref{fig:gaurav EVM}. Our third technique is to estimate the heart rates of subjects appearing in  videos by focusing on an area of facial skin, and analysing if any differences in heart rate calculations due to skin pulsation could be observed between them. 

The main hypothesis behind the work in this paper is to Euler magnify both  source and counterfeit videos in order to extract features which could highlight skin pulsation or minuscule movement by a subject in the videos. Through pulsation, we can  visualise and thus be able to extract the heart rate of the subject in the video by calculating the number of colour change peaks and counting each one as a heartbeat. 

EVM can amplify spatial as well as the temporal aspects of a video. Spatial magnifies the motion while temporal magnifies colour changes on skin tone. We use the temporal aspect to visualise the pulse on exposed facial skin. The features of this that are customisable are filter type, magnification factor and range of frequency.  Videos in our test set were subjected to a range of EVMs with a frequency range of between 1Hz (60 BPM) and 1.33 Hz (80 BPM). The amplification factor was set to 50. To fetch a heartbeat, a Fast Fourier transform algorithm was used. The temporal signal was transformed into a frequency domain to fetch the signal measured in hertz. 

Our data set consists of 19.25\% of  REAL videos which have not been altered, With the deepfakes accounting for 80.75\% of the samples. There is a huge imbalance between the categories thus models might be biased towards categorise videos as deepfake hence the data needs to be up-scaled to balance it.   

We used OpenCV \cite{bradski2008learning} to detect the locations of faces in videos using the face recognition package. We observe that in some cases, when the subject is not looking frontally at the camera or when the luminosity is low, the algorithm for face detection does not detect the face or eyes correctly.

\section{Results}
The results of our experiments  were produced  using Python 3  on a computer with 8 GB RAM and a 4 core Ryzen 5 AMD processor. We used 400 sets of videos from the DFDC Kaggle dataset \cite{dolhansky2019deepfake} and 30 assembled directly by us.
We generated 5 video datasets by changing parameters of Euler magnification, notably amplification factors. The complete  sequence can be divided into the following steps 
\begin{enumerate}[topsep=0pt,itemsep=-1ex,partopsep=1ex,parsep=1ex]
    \item Create metadata of the videos extracted from multiple sources;
    \item Detect faces and crop video to leave only the face in the video;
    \item Euler magnify the video with a specific frequency range and amplification factor;
    \item Train and evaluate a model on Euler magnified video.
\end{enumerate} 

\subsection{Results for Technique 1: SSIM}
As observed anecdotally from multiple graphs of SSIM scores for videos, there were more similarity score drops in counterfeit deepfake videos  when compared to their real video counterparts when magnified by the Euler magnification process. After magnifying the data using EVM, we calculated SSIM scores for all videos in the dataset. Deepfake video detection is particularly difficult to train as the manipulation can be observed only on a few frames and it is restricted to certain areas of the face. When there is much movement in the video, there can be inconsistencies and important areas in the frame appear only briefly.

Below are the results we obtained  when we used a range of standard  machine learning models to classify videos:

\begin{table}[ht]
\centering 
\begin{tabular}{lllll} 
\toprule
\multirow{2}{*}{\bf Technique} & {\bf Logistic} & {\bf Decision} & \multirow{2}{*}{\bf NNet} & {\bf NNet +}\\
& {\bf Regression} & {\bf Tree} & & {\bf LSTM} \\
\midrule 
Original Videos & 68.7\% & 65\% & 77\% & 77\% \\ 
EVM Videos & 53.7\% & 62\% & 70\% & 62\% \\ 
\bottomrule
\end{tabular}
\end{table}

These results are below the 82.56\% accuracy achieved by the best-performing submission to the DFDC but that was based on an ensemble while our results are one single technique. The results show consistently that SSIM-derived features are more discriminative of real vs. deepfake videos, before Euler Video Magnification was applied, suggesting that our default settings for EVM could be tweaked to improve performance.

In the table below, we processed our test videos with Euler magnification with  multiple amplification factors (15, 20, 30, 40 and 50). The frequency band was restricted to between 0.8hz and 1.0 Hz. As we can see from the results the lower amplification factor performed better than the higher amplification factor. In these videos, a higher amplification factor led to additional noise which blocked some of the features in the videos. This indicates that Euler magnification is introducing noise as the amplification factor increases. 

\begin{table}[ht]
\centering 
\begin{tabular}{lll}
\toprule
{\bf Amplification} & \multirow{2}{*}{\bf Accuracy} & \multirow{2}{*}{\bf Loss} \\
{\bf Factor} &&\\
\midrule
10 & 70.24\% & 0.6036\\ 
20 & 68.36\% & 0.6043\\ 
30 & 65.77\% & 0.6051\\ 
40 & 63.54\% & 0.6243\\ 
50 & 60.49\% & 0.6189\\ 
\bottomrule
\end{tabular}
\label{tab:SSIM}
\end{table}

\subsection{Results for Technique 2: LSTM}
We ran the Inception v3 inspired LSTM model  for 100 epochs to give the following results on our test video set. For these experiments we found that classification on the original unprocessed videos was more accurate and outperformed classification accuracy when EVM had been applied to the videos on an amplification factor of 30.  

\begin{table}[ht]
\centering 
\begin{tabular}{lll} 
\toprule
{\bf Video set} & {\bf Accuracy} & {\bf Loss} \\
\midrule 
Original Videos & 77.24\% & 0.88\\ 
EVM Videos & 61.79\% & 2.52\\ 
\bottomrule
\end{tabular}
\end{table}

\noindent 
This result is another apparent setback to the idea of using EVM  for deepfake detection. The difference in accuracy between original and EVM  videos is even more pronounced than in Model 1.

\subsection{Results for Technique 3: Heart Rate Estimation}

A comparison of heart rate based on pulsation observed in the Euler magnified video of an original and a deepfake video revealed that beats per minute (BPM) for both the videos were very similar, with slight changes only in the first decimal position.

\begin{figure}[htp]
    \centering
    \includegraphics[width=\columnwidth]{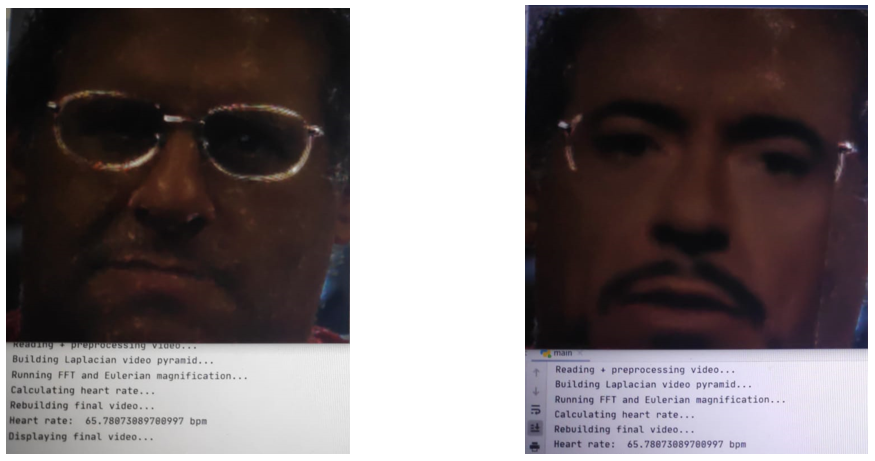}
    \caption{Heartbeat calculation using Fast Fourier transform on EVM of Original and Deepfake video}
    \label{fig:Heartbeat}
\end{figure}

As an example in Figure~\ref{fig:Heartbeat}, the heart rate of both the original and deepfake videos came to 65.78 beats per minute (1.096 Hz). Thus the temporal variance calculated through Euler video magnification is insufficient to differentiate deepfake from original videos \cite{bennett2016adaptive}.

Subjects appearing in both deepfake videos and original videos have an estimated heartbeat which is indistinguishable.
This reveals that the GANs used for creating the deepfakes does not simply superimpose a new image of the subject on top of the real image which could have concealed facial pulsation colour changes, but the GAN manages to faithfully model the true data distribution of the real data at the pixel level and in this way it keeps the colour-temporal changes from the genuine video, intact.

\section{Conclusions}

In this paper, we tested the effect and impact of Euler Video Magnification as a technique for video pre-processing leading to possible detection of deepfake videos. Both the colour and the spatial aspects of  EVM were tested as possibilities for a number of classification models we built for discriminating between real and deepfake. We used accuracy as a metric, even though accuracy is known not to be a great metric for evaluation when using imbalanced datasets like ours, which is why we include accuracy figures for both original and EVM processed videos, so we can compare.

What the accuracy performance figures do not show is that using EVM vs. not using EVM as a video pre-process actually detects {\em different} videos, so it is not the case that EVM pre-processing simply eliminates some  videos from being accurately classified as real or deepfake.  
Thus we should include a range of metrics including Precision, Recall, F1 and others so as to more fully understand what EVM is doing, but  as this paper is preliminary work, that forms part of future work. 

The results of the best performing systems at the Deepfake Detection Challenge \cite{dolhansky2019deepfake}  achieved an accuracy of 82.56\% across a much larger test dataset than we have used here, so we cannot compare our results directly against the data in the full DFDC results.  However because these best results were based on ensembles of techniques, it follows that the more variety among the systems used in the ensemble, the better will be will be the final, overall result.

We believe that our  results help us to understand how deepfake video generation incorporate pulsation information and subject tremor motion into their generated videos.  In future work, we would like to explore the most difficult facial objects to alter like lips and eyes, for fake detection as well as a deeper exploration into the results by using different evaluation metrics.  We would also investigate different feature sets and limit the parameters of EVM to focus on its impact.  We would also like to explore how compression artefacts in video storage interact with the EVM process as it is possible it is masking the contribution of EVM.

\subsection{Acknowledgements} 
This work was part-funded by Science Foundation Ireland through the Insight Centre for Data Analytics (SFI/12/RC/2289\_P2), co-funded by the European Regional Development Fund.


\begin{biography}
\noindent 
Rashmiranjan Das  works in the field of Artificial Intelligence at Deciphex Limited. He received his Masters in Data Analytics from Dublin City University in 2020 and holds a Bachelors in Computer Engineering from Mumbai University. After his undergraduate education, he joined Vertisystem limited as a software developer and then  as an Analyst at Ernst \& Young.
\\
\\
Gaurav Negi completed his Masters in Data Analytics from Dublin City University in 2020 and is now working  at Xcelerator Machine Translations Ltd. Gaurav holds a bachelor's degree in Computer Science from Amity University, India (2016) and worked for 3 years as a Data Analyst for Genpact India.
\\
\\
Alan Smeaton is Professor of Computing at Dublin City University. He received his PhD from University College Dublin (1987). He is an elected member of the Royal Irish Academy and a winner of the Academy's Gold Medal in Engineering Sciences, an award given  to individuals who have made a demonstrable and internationally recognised outstanding scholarly contribution in their fields.  Alan is Chair of ACM SIGMM.
\\
\end{biography}

\bibliographystyle{plain} 
\bibliography{EI-Paper.bib} 

\begin{thebibliography}{10}

\bibitem{abadi2016tensorflow}
Mart{\'\i}n Abadi, Paul Barham, Jianmin Chen, Zhifeng Chen, Andy Davis, Jeffrey
  Dean, Matthieu Devin, Sanjay Ghemawat, Geoffrey Irving, Michael Isard, et~al.
\newblock Tensorflow: A system for large-scale machine learning.
\newblock In {\em 12th USENIX Symposium on Operating Systems Design and
  Implementation}, pages 265--283, 2016.

\bibitem{arjovsky2017wasserstein}
Martin Arjovsky, Soumith Chintala, and L{\'e}on Bottou.
\newblock Wasserstein gan.
\newblock {\em arXiv preprint arXiv:1701.07875}, 2017.

\bibitem{bennett2016adaptive}
Stephanie~L Bennett, Rafik Goubran, and Frank Knoefel.
\newblock Adaptive eulerian video magnification methods to extract heart rate
  from thermal video.
\newblock In {\em 2016 IEEE International Symposium on Medical Measurements and
  Applications (MeMeA)}, pages 1--5. IEEE, 2016.

\bibitem{bourquard2019differential}
Aurélien Bourquard and Jeff Yan.
\newblock Differential imaging forensics.
\newblock {\em arXiv preprint arXiv:1906.05268}, 2019.

\bibitem{bradski2008learning}
Gary Bradski and Adrian Kaehler.
\newblock {\em {Learning OpenCV: Computer vision with the OpenCV library}}.
\newblock O'Reilly Media, Inc., 2008.

\bibitem{brophy2018interpretable}
Eoin Brophy, José Juan~Dominguez Veiga, Zhengwei Wang, Alan~F. Smeaton, and
  Tom{\'a}s~E. Ward.
\newblock An interpretable machine vision approach to human activity
  recognition using photoplethysmograph sensor data.
\newblock In {\em Proceedings of the 26th Irish Conference on Artificial
  Intelligence \& Cognitive Science \url{http://ceur-ws.org/Vol-2259/}}, pages
  244--255, December 2018.

\bibitem{chollet2015keras}
Fran\c{c}ois Chollet et~al.
\newblock Keras.
\newblock \url{https://keras.io}, 2015.
\newblock Last Accessed: 21 December 2020.

\bibitem{dolhansky2019deepfake}
Brian Dolhansky, Russ Howes, Ben Pflaum, Nicole Baram, and Cristian~Canton
  Ferrer.
\newblock The deepfake detection challenge ({DFDC}) preview dataset.
\newblock {\em arXiv preprint arXiv:1910.08854}, 2019.

\bibitem{fernandes2019predicting}
Steven Fernandes, Sunny Raj, Eddy Ortiz, Iustina Vintila, Margaret Salter,
  Gordana Urosevic, and Sumit Jha.
\newblock Predicting heart rate variations of deepfake videos using neural ode.
\newblock In {\em Proceedings of the IEEE International Conference on Computer
  Vision Workshops}, pages 0--0, 2019.

\bibitem{goodfellow2016nips}
Ian Goodfellow.
\newblock {NIPS} 2016 tutorial: Generative adversarial networks.
\newblock {\em arXiv preprint arXiv:1701.00160}, 2016.

\bibitem{guera2018deepfake}
David G{\"u}era and Edward~J Delp.
\newblock Deepfake video detection using recurrent neural networks.
\newblock In {\em 2018 15th IEEE International Conference on Advanced Video and
  Signal Based Surveillance (AVSS)}, pages 1--6. IEEE, 2018.

\bibitem{hochreiter1997lstm}
Sepp Hochreiter and J{\"u}rgen Schmidhuber.
\newblock {LSTM} can solve hard long time lag problems.
\newblock In {\em Advances in neural information processing systems}, pages
  473--479, 1997.

\bibitem{koopman2018detection}
Marissa Koopman, Andrea~Macarulla Rodriguez, and Zeno Geradts.
\newblock Detection of deepfake video manipulation.
\newblock In {\em The 20th Irish Machine Vision and Image Processing
  Conference}, pages 133--136, 2018.

\bibitem{kumar2020detecting}
Akash Kumar, Arnav Bhavsar, and Rajesh Verma.
\newblock Detecting deepfakes with metric learning.
\newblock In {\em 8th International Workshop on Biometrics and Forensics},
  pages 1--6. IEEE, 2020.

\bibitem{li2018detection}
Haodong Li, Bin Li, Shunquan Tan, and Jiwu Huang.
\newblock Detection of deep network generated images using disparities in color
  components.
\newblock {\em arXiv preprint arXiv:1808.07276}, 2018.

\bibitem{li2018ictu}
Yuezun Li, Ming-Ching Chang, and Siwei Lyu.
\newblock In ictu oculi: Exposing {AI} generated fake face videos by detecting
  eye blinking.
\newblock {\em arXiv preprint arXiv:1806.02877}, 2018.

\bibitem{maksutov2020methods}
Artem~A Maksutov, Viacheslav~O Morozov, Aleksander~A Lavrenov, and Alexander~S
  Smirnov.
\newblock Methods of deepfake detection based on machine learning.
\newblock In {\em IEEE Conference of Russian Young Researchers in Electrical
  and Electronic Engineering (EIConRus)}, pages 408--411. IEEE, 2020.

\bibitem{mccloskey2018detecting}
Scott McCloskey and Michael Albright.
\newblock Detecting {GAN}-generated imagery using color cues.
\newblock {\em arXiv preprint arXiv:1812.08247}, 2018.

\bibitem{mousavi2019blood}
Seyedeh~Somayyeh Mousavi, Mohammad Firouzmand, Mostafa Charmi, Mohammad
  Hemmati, Maryam Moghadam, and Yadollah Ghorbani.
\newblock Blood pressure estimation from appropriate and inappropriate {PPG}
  signals using a whole-based method.
\newblock {\em Biomedical Signal Processing and Control}, 47:196--206, 2019.

\bibitem{petrov2020deepfacelab}
Ivan Petrov, Daiheng Gao, Nikolay Chervoniy, Kunlin Liu, Sugasa Marangonda,
  Chris Um{\'e}, Jian Jiang, Luis RP, Sheng Zhang, Pingyu Wu, et~al.
\newblock Deepfacelab: A simple, flexible and extensible face swapping
  framework.
\newblock {\em arXiv preprint arXiv:2005.05535}, 2020.

\bibitem{singh2020using}
Simranjeet Singh, Rajneesh Sharma, and Alan~F. Smeaton.
\newblock Using {GANs} to synthesise minimum training data for deepfake
  generation.
\newblock In {\em Proceedings of the 28th Irish Conference on Artificial
  Intelligence \& Cognitive Science Dublin, Ireland
  \url{http://ceur-ws.org/Vol-2771/}}, pages 193--204, December 2020.

\bibitem{thies2016face2face}
Justus Thies, Michael Zollhofer, Marc Stamminger, Christian Theobalt, and
  Matthias Nie{\ss}ner.
\newblock Face2face: Real-time face capture and reenactment of {RGB} videos.
\newblock In {\em Proceedings of the IEEE conference on computer vision and
  pattern recognition}, pages 2387--2395, 2016.

\bibitem{wang2003multiscale}
Zhou Wang, Eero~P Simoncelli, and Alan~C Bovik.
\newblock Multiscale structural similarity for image quality assessment.
\newblock In {\em The 27th Asilomar Conference on Signals, Systems \&
  Computers}, volume~2, pages 1398--1402. IEEE, 2003.

\bibitem{wronski2019handheld}
Bartlomiej Wronski, Ignacio Garcia-Dorado, Manfred Ernst, Damien Kelly, Michael
  Krainin, Chia-Kai Liang, Marc Levoy, and Peyman Milanfar.
\newblock Handheld multi-frame super-resolution.
\newblock {\em ACM Transactions on Graphics}, 38(4):1--18, 2019.

\bibitem{wu2012eulerian}
Hao-Yu Wu, Michael Rubinstein, Eugene Shih, John Guttag, Fr{\'e}do Durand, and
  William Freeman.
\newblock Eulerian video magnification for revealing subtle changes in the
  world.
\newblock {\em ACM Transactions on Graphics (TOG)}, 31(4):1--8, 2012.

\end{thebibliography}

\end{document}